\pdfoutput=1
\documentclass[11pt]{article}

\usepackage{EMNLP2022}

\usepackage{times}
\usepackage{latexsym}

\usepackage[T1]{fontenc}
\usepackage[utf8]{inputenc}

\usepackage{microtype}

\usepackage{inconsolata}

\usepackage{soul}
\usepackage{url}
\usepackage{hyperref}
\usepackage{graphicx}
\usepackage{amsmath}
\usepackage{amsthm}
\usepackage{booktabs}
\usepackage{algorithm}
\usepackage{algorithmic}
\usepackage{arydshln}
\usepackage{subfigure}
\usepackage{multirow}

\title{Sentence Representation Learning with \\ Generative Objective rather than Contrastive Objective}

\author{Bohong Wu\textsuperscript{1,2}, Hai Zhao\textsuperscript{1,2,\thanks{*Corresponding author. This work was supported in part by the Key Projects of National Natural Science Foundation of China under Grants U1836222 and 61733011.}} \\
    	$^{1}$ Department of Computer Science and Engineering, Shanghai Jiao Tong University \\
    	$^{2}$ Key Laboratory of Shanghai Education Commission for Intelligent Interaction \\
and Cognitive Engineering, Shanghai Jiao Tong University \\	\texttt{chengzhipanpan@sjtu.edu.cn,zhaohai@cs.sjtu.edu.cn} \\}

\begin{document}
\maketitle
\begin{abstract}
Though offering amazing contextualized token-level representations, current pre-trained language models take less attention on accurately acquiring sentence-level representation during their self-supervised pre-training. However, contrastive objectives which dominate the current sentence representation learning bring little linguistic interpretability and no performance guarantee on downstream semantic tasks. We instead propose a novel generative self-supervised learning objective based on phrase reconstruction. To overcome the drawbacks of previous generative methods, we carefully model intra-sentence structure by breaking down one sentence into pieces of important phrases. Empirical studies show that our generative learning achieves powerful enough performance improvement and outperforms the current state-of-the-art contrastive methods not only on the STS benchmarks, but also on downstream semantic retrieval and reranking tasks. Our code is available at \href{https://github.com/chengzhipanpan/PaSeR}{https://github.com/chengzhipanpan/PaSeR}.
\end{abstract}

\section{Introduction}

Sentence Representation Learning has long been a hot research topic \cite{conneau2017supervised,cer2018universal}, for its effectiveness in a variety of downstream tasks like information retrieval and question answering \cite{yang2018hotpotqa}.

Although pre-trained language models (PrLMs) like BERT \cite{devlin2019bert} have achieved overwhelming performance on various token-level tasks, they are also criticized for being unable to produce high-quality sentence-level representations. Research \cite{li2020emnlp} has shown that the native sentence representation produced by the “[CLS]” token of BERT shows extremely poor performance on sentence evaluation benchmarks like semantic textual similarity (STS) tasks.

The primary cause of these low-quality sentence representations is the lack of effective self-supervised sentence-level training objectives. As discussed in ConSERT \cite{yan-etal-2021-consert}, the original sentence-level pretraining objective Next Sentence Prediction (NSP) is too weak to provide high-quality sentence representation. Therefore, recent researchers are seeking other effective self-supervised sentence-level objectives \cite{carlsson2020semantic,yan-etal-2021-consert,gao2021simcse}. Generally, self-supervised methods include both (i) generative methods, like Masked Language Modelling (MLM), and (ii) contrastive methods, like Next Sentence Prediction (NSP). 

% Generative methods这里的介绍，感觉上有欠缺，不够好。要多想想看。

By treating one sentence as a whole and contrasting sentence representations with each other in the same training batch, contrastive methods have been shown extremely effective in Sentence Representation Learning in recent years. Generally, contrastive methods often use various data augmentation techniques to create different views for one sentence, and align the representations of these views within the same batch. ConSERT \cite{yan-etal-2021-consert} utilizes techniques including token shuffling, feature cutoff, etc., and provides a general contrastive learning framework for Sentence Representation Learning. SimCSE \cite{gao2021simcse} suggests using different dropout masks is a simple yet more powerful augmentation technique for creating different views of one sentence. 

Though effective, there also exist several drawbacks in contrastive methods. (i) The training procedure of contrastive methods lacks enough linguistic interpretability. What information is encoded into the sentence representation is kept unknown. (ii) As suggested in TSDAE \cite{wang-etal-2021-tsdae-using}, good performance on STS tasks by contrastive methods does not ensure good performance on downstream tasks like semantic retrieval and reranking because of the obvious inductive bias. 

% (ii) Experiments show that when combined with MLM objective, SimCSE suffers from significant performance degradation, indicating that the high-quality sentence representation of SimCSE is at the cost of sacrificing the quality of token representations, which limits its application scenarios where both qualities are valued. 

On the contrary, generative self-supervised learning techniques offer researchers good interpretability and controllable training by enabling the choice of what to generate. However, although generative methods like MLM have achieved overwhelming performance in token representation learning, little effort has been put into investigating the potential of generative methods in sentence representation learning. Cross-Thought \cite{wang2020cross} and CMLM \cite{yang2021universal} are the most representative generative methods, which both leverage the contextual sentence representations to recover the masked tokens in one sentence. Unfortunately, they highly depend on the contextual information, and mainly focus on the document-level corpus, thus performing unsatisfyingly in STS tasks where representations of short texts are valued. Latter, TSDAE \cite{wang-etal-2021-tsdae-using} proposes to use a de-noising auto-encoder to recover the original sentence from the corrupted version. Although TSDAE does not depend on document-level corpus anymore, it still suffers from inferior performance on the STS benchmarks.

\begin{figure}
    \centering
    \includegraphics[width=0.9\linewidth]{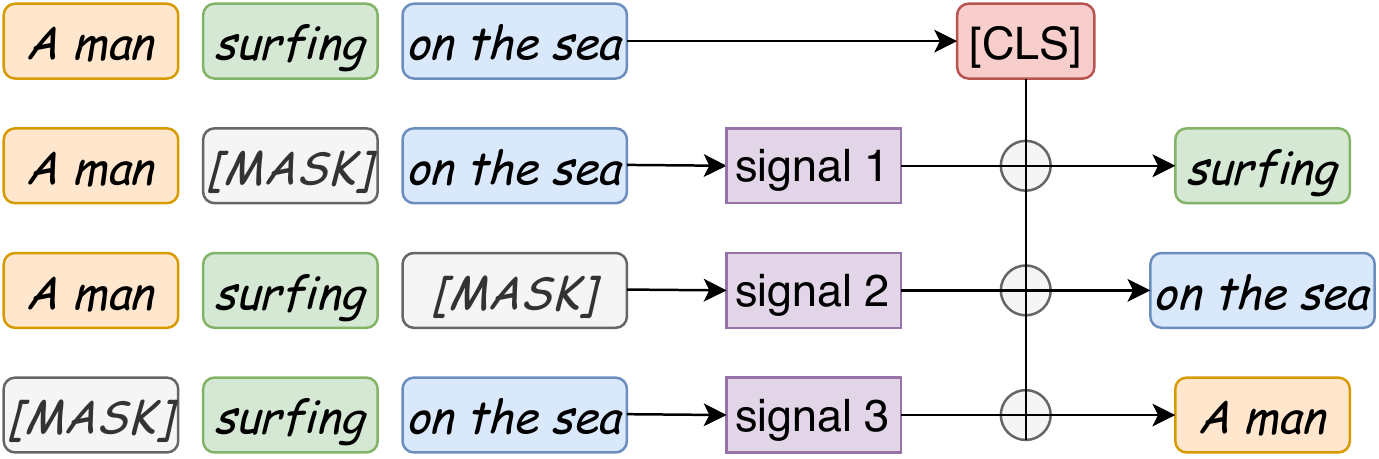}
    \caption{A description of the design intuition of our PaSeR.}
    %After encoding, the sentence representation should be able to recover the important phrases of the original sentence when given appropriate signals.}
    \label{fig:sketchy}
\end{figure}

We attribute the inferior performance of existing generative methods to the wrong modeling perspective. The existing generative methods still follow the training paradigm of contrastive methods, which considers one sentence as an inseparable whole, and learn sentence representations from the \textbf{inter-sentence perspective}. In this paper, we novelly propose to model sentence representation learning from the \textbf{intra-sentence perspective}. We especially emphasize the importance of the semantic components within the sentence, i.e. phrases. 
We therefore present \textbf{P}hrases-\textbf{a}ware \textbf{Se}ntence \textbf{R}epresentation (PaSeR), which explicitly encodes the representation of the most important phrases into sentence representations.
In detail, we hypothesize a good sentence representation should be able to encode and reconstruct the important phrases in the sentence when given a suitable generation signal. 
Inspired by SimCSE \cite{gao2021simcse}, we provide such generation signals by our \textit{duplication and masking} strategy. As shown in Figure \ref{fig:sketchy}, we mask out important phrases in the original sentences, and encode these masked sentences to provide signals for phrase reconstruction. Experiments show that our PaSeR achieves the SOTA performance on multiple single tasks in STS in both unsupervised and supervised settings, and especially better average STS performance than SimCSE in the supervised setting. Extensive experiments further present that our PaSeR achieves better performances on downstream tasks including semantic retrieval and reranking.

% (iii) Contrastive loss is incompatible with the original token-level pre-training objective MLM. According to SimCSE, the performance of SimCSE drops drastically on the STS tasks when the contrastive loss is combined with the MLM loss, indicating that the contrastive objective sacrifices the quality of token-level representation.

% In this paper, we investigate the potential of generative self-supervised techniques in Sentence Representation Learning. Unlike previous researchers who mainly study the inter-sentence relationship, we focus more on the intra-sentence relationship, and emphasize the importance of the semantic components within the sentence, i.e. phrases. We present \textbf{P}hrases-\textbf{a}ware \textbf{Se}ntence \textbf{R}epresentation (PaSeR), which explicitly encodes the representation of each phrase into sentence representations. In detail, we hypothesize a good sentence representation should be able to reconstruct the important phrases in the sentence when given a suitable generation signal. Inspired by \cite{gao2021simcse}, we provide such generation signals by a \textit{duplication and masking} strategy. Concretely, we mask out several phrases in the original sentences, and encode these masked sentences to provide hints for phrase reconstruction, depicted in Figure \ref{fig:sketchy}. Experiments show that our PaSeR achieves the SOTA performance on multiple single tasks in STS, and also comparable average performance in STS with the current SOTA contrastive method SimCSE in the unsupervised setting. Extensive experiments further present that our PaSeR achieves better semantic retrieval performance than SimCSE on the Quora Question Pair dataset.

\noindent\textbf{Contributions} (i) We propose an effective generative self-supervised objective of training sentence representations without leveraging document-level corpus. Based on such an objective we present PaSeR, a Phrase-aware Sentence Representation Learning model. (ii) Experiments show that our proposed PaSeR achieves SOTA performance on multiple single tasks in STS, and especially better average STS performance than previous best contrastive method, SimCSE, in the supervised setting. Our PaSeR provides an effective alternative for Sentence Representation Learning against the current trend of contrastive methods.

% 其实就是现在一些工作

% 如何吧related work搞成
\section{Related Work}

\subsection{Supervised Sentence Representations}
Supervised sentence representations leverage the idea of transfer learning. Previous works \cite{conneau2017supervised,cer2018universal} have shown that utilizing labeled datasets from Natural Language Inference (NLI) is extremely helpful for Sentence Representation Learning. Based on these researches, Sentence-BERT \cite{reimers2019sentence} introduces siamese BERT encoders with shared parameters and trains them on NLI datasets, achieving acceptable performance on STS tasks. Although these supervised methods can provide high-quality sentence representations, the labeling cost of sentence pairs still urges the researchers to search for a more effective unsupervised solution.
% to handle other languages where labeled sentence pairs are rare.

\subsection{Post-processing of BERT Representations}
Several post-processing methods are first proposed to improve the sentence representations produced by original BERT. Generally, these methods analyze the distorted sentence representation space, and propose changing the representation space to isotropic Gaussian ones via flow methods like BERT-flow \cite{li2020emnlp} or simple projection methods like BERT-whitening \cite{su2021whitening}. However, their performance is very limited, as their sentence representations are not finetuned due to the lack of suitable sentence-level training objectives in the original BERT model. 

\subsection{Self-supervised Sentence-level Pre-training}
Recently, researchers are seeking more effective sentence-level pre-training objectives, from the aspects of both generative ones and contrastive ones.

\noindent\textbf{Generative Methods} Little efforts have been paid into studying what generation methods can achieve in Sentence Representation Learning. Among these works, Cross-thought \cite{wang2020cross} and CMLM \cite{yang2021universal} are the most representative ones, which both propose to recover masked tokens of one sentence by the contextual-sentence representations. However, in both methods, document-level training data are needed, making it unsuitable for evaluating the similarity between short texts. Recently TSDAE \cite{wang-etal-2021-tsdae-using} also present a generative method, which aims to recover the original sentence from a corrupted version. Although TSDAE doesn't need contextual texts anymore, it suffers from inferior performance on the STS benchmarks. 

\noindent\textbf{Contrastive Methods} Recently, contrastive learning has presented its superiority in Sentence Representation Learning. Generally, existing contrastive methods are seeking effective ways of creating different views of one sentence, pushing their representations closer while pulling views of different sentences away. Contrastive Tension (CT) \cite{carlsson2020semantic} introduces the Siamese network structure to create different views of one sentence, and treat different views from one sentence as positive pairs while others as negative pairs. ConSERT \cite{yan-etal-2021-consert} creates different views of sentences by data augmentation techniques including token shuffling, feature cutoff, and adversarial attacks. After that, SimCSE \cite{gao2021simcse} presents that the original dropout mask design is already a very effective data augmentation strategy, and has achieved the SOTA performance on the STS benchmarks.

\section{Method}

% In this section, we introduce the detail of our \textbf{P}hrase-\textbf{a}ware \textbf{Se}ntence \textbf{R}epresentation (PaSeR).

% 这个图后面肯定要换一个，目前比较垃圾。至少要让他占半夜把？加点料，不要那么像SIMCSE

\subsection{Data Pre-processing}

\subsubsection{Phrase Extraction}

Phrase extraction is the core component of our PaSeR. Which phrase to generate directly determines what information we encode in the sentence representation. In this paper, we mainly use two off-the-shelf methods to extract the important phrases. 

\noindent$\bullet$ A random sub-tree from the syntax parsing tree of one sentence. By using NLTK \cite{bird-loper-2004-nltk}, we can easily extract components of one sentence including Subordinate Clauses (SBAR), Verb Phrases (VP), and Noun Phrases (NP). 

\noindent$\bullet$ A statistical key-phrase extraction algorithm, RAKE \cite{rose2010automatic}. RAKE first tokenizes the original sentence or document into phrases using stopwords or punctuations. To acquire the importance of each phrase, RAKE first constructs the co-occurrence matrix of words, and computes the importance of word $w$ via its degree $deg(w)$ in the co-occurrence matrix and its frequency $freq(w)$ in the document or sentence by $$wordScore(w) = deg(w) / freq(w)$$ Finally, the importance of one phrase is calculated by summation of $wordScore$ on all words in the phrase.

Experimentally, the second method achieves much better performance on the STS tasks. Therefore, if not specified, we use the second method for phrase extraction. We will discuss the performance difference between these two extraction methods in Section \ref{sec:phrase_to_mask}. 

\begin{figure*}[tp]
    \centering
    \includegraphics[width=0.95\linewidth]{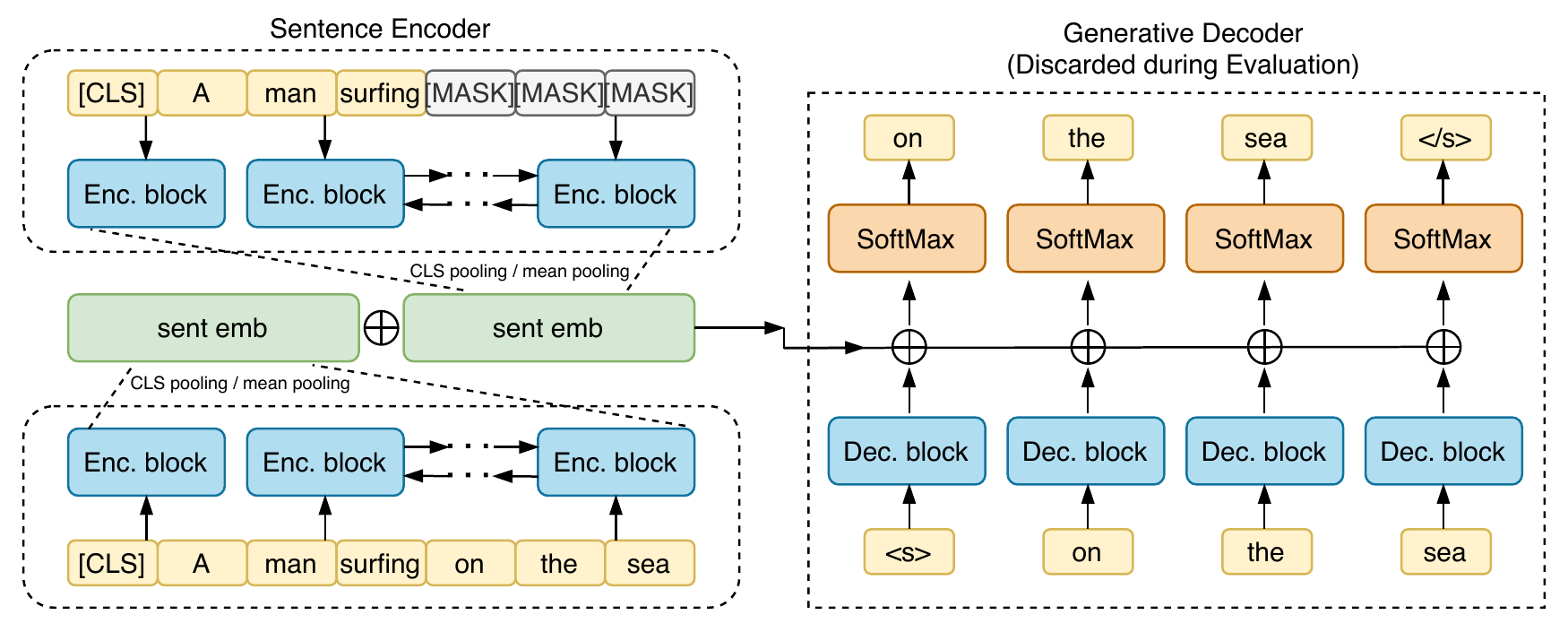}
    \caption{Overview of PaSeR. The left part presents the \textit{Sentence Encoders} where parameters are shared. The right part presents the \textit{Generative Decoder}, which will be discarded during evaluation. Therefore, the extra parameters are only used during the training stage, and do not affect the inference speed at the evaluation stage.}
    \label{fig:paser_pipe}
\end{figure*}

\subsubsection{Duplicate and Masking}
For the motivation of recovering important phrases within one sentence, we propose to use difference modeling. In detail, for one given sentence $s$, which is composed of multiple phrases $\mathcal{P} = \{p_0, p_1, ..., p_n\}$, ordered by their importance calculated by RAKE \cite{rose2010automatic}. To recover the most important phrase like $p_0$, it is natural to come up with the following equality:
\begin{equation}
 p_0 = \mathcal{P} - \mathcal{P}/\{p_0\}  
\end{equation}

Therefore, we duplicate such $s$ as $\Tilde{s}$, but mask out the most important phrase $p_{0}$ that we need to generate, shown in the left part of Figure \ref{fig:paser_pipe}. Denote the sentence encoder as $Enc$, we can get the sentence representation of $E_s = f_{Enc}(s)$ and $E_{\Tilde{s}} = f_{Enc}(\Tilde{s})$. By combining the representations of both $E_{s}$ and $E_{\Tilde{s}}$, we can recover the masked phrases $p_0$ with a suitable transformer decoder.

\subsubsection{Data Augmentation}

To improve the robustness of the sentence representations produced by our PaSeR, following EDA \cite{wei-zou-2019-eda}, we introduce data augmentation on both $s$ and $\Tilde{s}$ before the paired sentences are fed into the sentence encoder. We mainly use three types of data augmentation strategies including \textit{Synonym Replacement}, \textit{Random Deletion} and \textit{Token Reordering}. 

We speculate that, (i) Using \textit{Synonym Replacement} on both $s$ and $\Tilde{s}$ is an effective strategy to create semantic similar phrases with different tokens, which helps the model capture the semantic similarities instead of token similarities. (ii) \textit{Random Deletion} strategy can well alleviate the effect brought by frequent words or phrases. (iii) \textit{Token Reordering} strategy can make our sentence encoder less sensitive to token orders and changes in positional embeddings.

% Therefore, after encoding $\Tilde{s}$ as $E_{\Tilde{s}}$, $E_{\Tilde{s}}$ can be viewed as the guiding signal for generation of the missing phrase $p_{i}$. 

\subsection{Unsupervised PaSeR}
% 这里就可以讲讲我们两种masking的形式。不然method还真不好讲
\subsubsection{Sentence Encoder} Following previous works \cite{li2020emnlp,su2021whitening}, our sentence encoders are based on the pretrained language model, BERT \cite{devlin2019bert}. The pooling methods include (i) directly using the "[CLS]" token representation, (ii) averaging the token representations in the last layer of BERT, (iii) using a weighted average of token representations from the intermediate layers of BERT, and we choose the best pooling method based on its performance on the STS tasks.

\subsubsection{Decoding Signal} After acquiring the sentence representation of both $E_{s}$ and $E_{\Tilde{s}}$, the way of combining these two representations also plays an important role in sentence representation learning. Following SBERT \cite{reimers2019sentence} but a step further, we use a weighted combination of $E_s, E_{\Tilde{s}}, |E_s - E_{\Tilde{s}}|, E_s * E_{\Tilde{s}}$ to create the decoding signal for the following generative decoder:
\begin{equation}
Signal_{Dec} = [E_s, E_{\Tilde{s}}, m*|E_s - E_{\Tilde{s}}|, n*|E_s * E_{\Tilde{s}}|]
\end{equation}

Here, $m$ and $n$ are scaling factors to normalize these four decoding signals, and both variables are selected by grid search. We will discuss the selection of $m$ and $n$ in Appendix \ref{sec:dec_signal}.

\subsubsection{Generative Decoder}
The generative decoder performs as a representation regularizer for training the sentence encoder, and can be discarded during the evaluation stage. Therefore, the decoder does not add up any additional hyperparameters for downstream tasks. In our experiments, we use variants of Transformer \cite{vaswani2017attention} decoders as our phrase reconstruction decoder $Dec$. Suppose now the masked phrase $p_0$ is composed of several tokens $\{t_1, t_2, ..., t_k\}$, and given the decoding signal $Signal_{Dec}$, the phrase reconstruction process is formulated as:
\begin{equation}
L_{generative} = -\sum_{i=1}^{k} \log P_{f_{Dec}}\left(t_{i} \mid t_{<i}, Signal_{Dec}\right)    
\end{equation}

\subsubsection{Combined with MLM}
To preserve the quality of token-level representation, we also incorporate the MLM objective with our reconstruction objective. The final training loss is a combination:
\begin{equation}
L_{total} = L_{MLM} + L_{generative}
\end{equation}

% 我们还没有tune这个parameter，说不定这个就是超越76的关键呢。不过不是很看好

\subsection{Supervised PaSeR}

For supervised settings, our PaSeR loss design can be easily incorporated with the frontier supervised methods. Moreover, our unsupervised PaSeR can provide a good initializing checkpoint for training the supervised sentence encoder.

\begin{figure}
    \centering
    \includegraphics[width=1.0\linewidth]{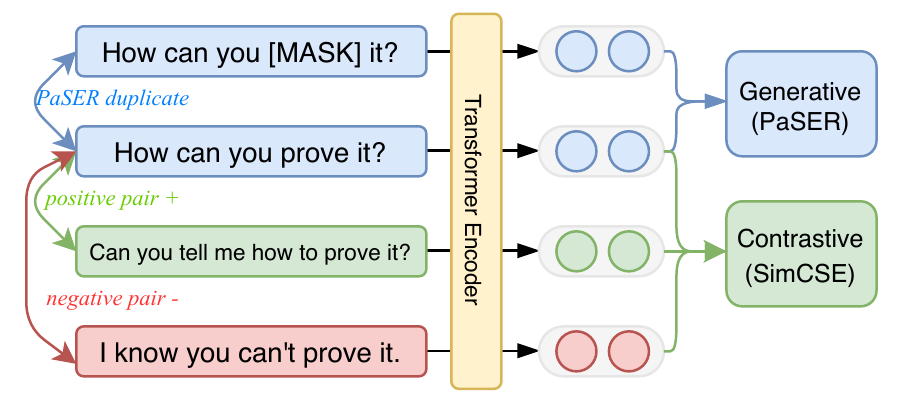}
    \caption{PaSeR baseline in the supervised setting.}
    \label{fig:super_baseline}
\end{figure}

Generally, methods of incorporating supervised signals in sentence representation learning can be divided into two types. (i) A sequence classification training objective following SBERT \cite{reimers2019sentence}. (ii) A contrastive learning objective following SimCSE \cite{gao2021simcse}. Given the prominent performance of the latter approach, we combined the contrastive loss introduced in SimCSE with our generative PaSeR loss. The training process is shown in Figure \ref{fig:super_baseline}. We initialize the sentence encoder from the best checkpoint of our unsupervised PaSeR. The final loss function is formulated as:
\begin{equation}
L_{supervise} = L_{contrastive} + \alpha L_{generative}    
\end{equation}
where $\alpha$ is an adjustable hyper-parameter that is searched in our experiments.

\begin{table*}[tp]
    \centering
    \setlength{\tabcolsep}{6pt}
    \fontsize{10.0pt}{\baselineskip}\selectfont
    \begin{tabular}{lcccccccc}
    \toprule
    Model & STS12 & STS13 & STS14 & STS15 & STS16 & STS-B & SICK-R & Avg. \\
    \midrule
    % \multicolumn{9}{l}{\textit{Frontier Unsupervised Methods}} \\
    \multicolumn{9}{l}{\textit{Original BERT/Glove}} \\
    GloVe embeddings & 55.14 & 70.66 & 59.73 & 68.25 & 63.66 & 58.02 & 53.76 & 61.32 \\
    BERT$_{base}$-[CLS] & 21.54 & 32.11 & 21.28 & 37.89 & 44.24 & 20.29 & 42.42 & 31.40 \\
    BERT$_{base}$-mean & 30.87 & 59.89 & 47.73 & 60.29 & 63.73 & 47.29 & 58.22 & 52.57 \\
    BERT$_{base}$-first-last avg. & 39.70 & 59.38 & 49.67 & 66.03 & 66.19 & 53.87 & 62.06 & 56.70 \\
    
    \hdashline
    
    \multicolumn{9}{l}{\textit{Post-Processing Methods}} \\
    BERT-flow & 58.40 & 67.10 & 60.85 & 75.16 & 71.22 & 68.66 & 64.47 & 66.55 \\
    BERT-whitening & 57.83 & 66.90 & 60.90 & 75.08 & 71.31 & 68.24 & 63.73 & 66.28\\
    \hdashline
    
    \hdashline
    
    \multicolumn{9}{l}{\textit{Contrastive Methods}} \\
    % CLEAR$_{base}$ & 49.0 & 48.9 & 57.4 & 63.6 & 65.6 & 75.6 & 72.5 & 61.8 \\
    CT-BERT$_{base}$ & 66.86 & 70.91 & 72.37 & 78.55 & \underline{77.78} & - & - & - \\
    ConSERT$_{base}$ & 64.64 & 78.49 & 69.07 & 79.72 & 75.95 & 73.97 & 67.31 & 72.74 \\
    SimCSE-BERT$_{base}$ + MLM & 44.40 & 66.60 & 51.27 & 67.48 & 67.95 & 52.44 & 59.86 & 58.57 \\
    SimCSE-BERT$_{base}$ & 68.40 & \underline{82.41} & \textbf{74.38} & \underline{80.91} & \textbf{78.56} & 76.85 & 72.23 & \textbf{76.25} \\ 

    \hdashline
    \multicolumn{9}{l}{\textit{Generative Methods}} \\
    % IS-BERT$_{base}$ & 56.77 & 69.24 & 61.21 & 75.23 & 70.16 & 69.21 & 64.25 & 66.58 \\
    CMLM-BERT$_{base}$ & 58.20 & 61.07 &  61.67 &  73.32 & 74.88 & 76.60 & 64.80 & 67.22 \\
    
    PaSeR-BERT$_{base}$ (Our Method) & \textbf{70.21} & \textbf{83.88} & \underline{73.06} & \textbf{83.87} & 77.60 & \textbf{79.19} & 65.31 & \underline{76.16} \\
    \bottomrule
    \end{tabular}
    \caption{Unsupervised sentence representation performance on STS tasks. Bold statistics represent the best performance while underlined ones represent the second-best performance.}
    \label{tab:main_exp}
\end{table*}

\section{Experiment}

\subsection{Evaluation Datasets}

% \subsubsection{Semantic Similarity} 

\noindent\textbf{Semantic Textual Similarity (STS)} For the similarity evaluation of sentence representations, following previous works \cite{su2021whitening,gao2021simcse,yan-etal-2021-consert}, we use the Semantic Textual Similarity (STS) datasets as our evaluation benchmark, including STS tasks
2012-2016 (STS12-STS16) \cite{agirre2012semeval,agirre2013sem,agirre2014semeval,agirre2015semeval,agirre2016semeval}, STS Benchmark (STS-B) \cite{cer2017semeval} and SICK-Relatedness (SICK-R) \cite{marelli2014sick}. Samples in datasets are paired sentences with human-labeled relatedness scores from 0 to 5. We use the spearman correlation x 100 on these 7 STS datasets to evaluate and compare the performance between all baselines and frontier researches.

% \subsubsection{Semantic Retrieval/Reranking} For the downstream semantic retrieval task, we use the following two datasets for evaluation. 

\noindent\textbf{Quora Question Pair} Quora Question Pair dataset \footnote{\url{https://quoradata.quora.com/First-Quora-Dataset-Release-Question-Pairs}} (QQP) consists of over 400,000 lines of potential question duplicate pairs, denoted as $(q_1, q_2)$. We collect all the $q_2$s as the question corpus, and all $q_1$s that have at least a positive paired $q_2$ as the query set. We then use the query set to retrieve similar questions from the question corpus. The evaluation metrics contain Mean Average Precision (MAP) and Mean Reciprocal Rank (MRR). 

\noindent\textbf{AskUbuntu Question} AskUbuntu Question dataset is a semantic reranking dataset, which contains a pre-processed collection of questions taken from AskUbuntu.com\footnote{\url{https://askubuntu.com/}} 2014 corpus dump. Different from QQP, the question corpus for each query is given with the size of 20, and models are required to re-rank these 20 given questions according to the similarity measurement. We also use MAP and MRR as evaluation metrics. 

\subsection{Training Details}

We use BERT-base \cite{devlin2019bert} as the sentence encoder for all experiments. Following SimCSE \cite{gao2021simcse}, we restrict the maximum sequence length to 32 with an initial learning rate of 3e-5. The batch size is selected from [32, 64, 96]. For training in the unsupervised setting, following ConSERT \cite{yan-etal-2021-consert}, we mix all the unlabeled texts from the seven STS datasets as the training data. For training in the supervised setting, given the superior performance of SimCSE \cite{gao2021simcse}, we also use a combination of SNLI dataset \cite{bowman-etal-2015-large} and MNLI dataset \cite{williams-etal-2018-broad}, and combine the contrastive learning framework of SimCSE with our PaSeR to train the supervised sentence encoders.

Following previous works \cite{yan-etal-2021-consert,gao2021simcse}, we use the development set of STS-B to choose the best-performing model. If not specified, we take the "[CLS]" representation as the sentence representation for most of the experiments, and discuss different effects when different pooling methods are adopted in Appendix \ref{sec:pooling_method}.

For the generative decoder, any type of transformer \cite{vaswani2017attention} decoder is acceptable. If not specified, we use a 6-layer transformer decoder as the generative decoder, and the word embedding layer is shared between the sentence encoder and the generative decoder. We also present a discussion on how the complexity of the decoder affects the performance of the sentence encoder in Section \ref{discuss:sent_encoder}. The weights of the generative decoder are randomly initialized.

\begin{table*}[tp]
    \centering
    \setlength{\tabcolsep}{6pt}
    \fontsize{10.0pt}{\baselineskip}\selectfont
    \begin{tabular}{lcccccccc}
    \toprule
    Model & STS12 & STS13 & STS14 & STS15 & STS16 & STS-B & SICK-R & Avg. \\
    \midrule
    
    % \multicolumn{9}{l}{\textit{Frontier Supervised Methods}} \\
    Universal Sentence Encoder & 64.49 & 67.80 & 64.61 & 76.83 & 73.18 & 74.92 & 76.69 & 71.22 \\
    SBERT$_{base}$ & 70.97 & 76.53 & 73.19 & 79.09 & 74.30 & 77.03 & 72.91 & 74.89 \\
    SBERT$_{base}$-flow & 69.78 & 77.27 & 74.35 & 82.01 & 77.46 & 79.12 & 76.21 & 76.60 \\
    SBERT$_{base}$-whitening & 69.65 & 77.57 & 74.66 & 82.27 & 78.39 & 79.52 & 76.91 & 77.00 \\
    CT-SBERT$_{base}$ & 74.84 & 83.20 & 78.07 & 83.84 & 77.93 & 81.46 & 76.42 & 79.39 \\
    % \multicolumn{9}{l}{\textit{Contrastive Methods}} \\
    ConSERT-BERT$_{base}$ & 74.07 & 83.93 & 77.05 & 83.66 & 78.76 & 81.36 & 76.77 & 79.37 \\
    SimCSE-BERT$_{base}$ & 75.30 & 84.67 & 80.19 & 85.40 & 80.82 & \textbf{84.25} & \textbf{80.39} & 81.57 \\ 
    
    % \hdashline
    % \multicolumn{9}{l}{\textit{Generative Methods}} \\
    \hdashline
    % \multicolumn{9}{l}{\textit{Variants of Our Methods}} \\
    % PaSeR-BERT$_{base}$-single & - & - & - & - & - & - & - & - \\
    PaSeR-BERT$_{base}$ (Our Method) & \textbf{78.40} & \textbf{85.80} & \textbf{80.79} & \textbf{86.36} & \textbf{82.95} & 83.67 & 80.31 & \textbf{82.61} \\
    
    \bottomrule
    \end{tabular}
    \caption{Supervised sentence representation performance on STS tasks. Bold statistics represent the best performance among all baselines.}
    \label{tab:supervise_exp}
\end{table*}
% We will discuss the performance difference when choices of decoders are varied in section \ref{abl:decoder}.

\subsection{Results on Semantic Textual Similarity}
% can we achieve better results with contrastive learning?
% 对于图片来说可能
% 目前看了下，换组数据都能有很好的sickr。sickr应该是很好搞的。
\noindent\textbf{Unsupervised Settings} The performance of our PaSeR and other frontier researches are presented in Table \ref{tab:main_exp}. Here, we separate all these frontier researches into four categories. (i) Original baselines including different pooling methods of BERT and Glove embeddings. (ii) Post-processing baselines including BERT-flow \cite{li2020emnlp} and BERT-whitening \cite{su2021whitening}. (iii) Contrastive methods including CT-BERT$_{base}$ \cite{carlsson2020semantic}, ConSERT \cite{yan-etal-2021-consert} and SimCSE \cite{gao2021simcse}. (iv) Generative methods including CMLM \cite{yang2021universal} and our PaSeR.

% (iii) Supervised baselines include Universal Sentence Encoder (USE) \cite{conneau2017supervised} and Sentence BERT \cite{reimers2019sentence} which is finetuned on the NLI datasets.

% ------- Augmentation Ablation Table ---------
% \begin{table*}[tp]
%     \centering
%     \begin{tabular}{lcccccccc}
%     \toprule
%     Model & STS12 & STS13 & STS14 & STS15 & STS16 & STS-B & SICK-R & Avg. \\
%     \midrule
%     BERT$_{base}$-[CLS] & 21.54 & 32.11 & 21.28 & 37.89 & 44.24 & 20.29 & 42.42 & 31.40 \\
    
%     PaSeR-no aug & 63.25 & 79.66 & 70.70 & 80.21 & 75.35 & 74.47 & 62.45 & 72.30 \\
    
%     \ + Random Swapping & 65.21 & 79.64 & 69.45 & 80.53 & 76.11 & 74.98 & 63.45 & 72.77 \\
    
%     \ + Random Deletion & 65.62 & 79.38 & 68.75 & 80.77 & 76.61 & 75.49 & 63.79 & 72.92 \\
    
%     \ + Synonymy Replacement & \textbf{71.09} & \textbf{83.54} & \textbf{72.58} & \textbf{83.49} & \textbf{77.12} & \textbf{78.35} & \textbf{65.02} & \textbf{75.88} \\
%     \bottomrule
%     \end{tabular}
%     \caption{Ablation study of data augmentation strategies.}
%     \label{tab:data_aug}
% \end{table*}

From Table \ref{tab:main_exp}, (i) Under the unsupervised setting, our PaSeR achieves the SOTA performance on several benchmark datasets like STS12, STS13, STS15, STS-B, and is also the second-best model considering the average performance. (ii) Unlike our PaSeR which is naturally combined with MLM, SimCSE suffers from significant performance degradation on the STS benchmarks when combined with the MLM objective. (iii) Compared to the previous best generative method CMLM, PaSeR achieves an average of 8.94 absolute performance gain. Such improvement especially presents the superiority of the intra-sentence modeling perspective within the scope of generative methods.

% (ii) Performance of SimCSE drops drastically when combined with the MLM objective, but our PaSeR is naturally incorporated with MLM. We hypothesize this is because the contrastive learning objective has a very different convergence speed from the MLM objective, making the model difficult to balance between the quality of token-level representation and sentence-level representation.

% comparable performance to SimCSE, while better performance than ConSERT in all directions.

\noindent\textbf{Supervised Settings} The results of supervised sentence encoders are shown in Table \ref{tab:supervise_exp}.
By initiating the sentence encoder from a previous best unsupervised checkpoint, our PaSER can achieve an average of 1.04 performance gain on the STS benchmark, compared with the SimCSE baseline. The results demonstrate that using our PaSeR design is an effective self-supervised learning objective that could provide high-quality sentence representation during the pre-training stage.

\subsection{Results on Semantic Retrieval/Reranking}
\begin{table}[tp]
	\centering
	\setlength{\tabcolsep}{1.5pt}
	\fontsize{10.0pt}{\baselineskip}\selectfont
	\begin{tabular}{lcccc}
		\toprule
		\multirow{2}{*}{Method} & \multicolumn{2}{c}{QQP} & \multicolumn{2}{c}{AskUbuntu} \\
		& MAP & MRR$_{@10}$ & MAP & MRR$_{@10}$ \\
		\midrule
		\multicolumn{5}{l}{\textit{Unsupervised Baselines}} \\
		BERT$_{base}$ & 72.15 & 32.50 & 41.51 & 53.80 \\
		
		BERT-whitening$_{base}$ & 74.67 & 61.43 & 46.15 & 60.11 \\
		
% 		BERT-whitening$_{large}$ & 61.94 & 62.28 & - & - \\
		
		SimCSE-BERT$_{base}$ & 75.42 & 62.69 & \textbf{50.63} & 63.49 \\
		
		PaSeR-BERT$_{base}$ & \textbf{76.30} & \textbf{65.07} & 50.51 & \textbf{64.56} \\
		
		\hdashline
		\multicolumn{5}{l}{\textit{Supervised Baselines}} \\
		SimCSE-BERT$_{base}$ & 76.17 & 63.94 & 51.50 & 66.16 \\
		
		PaSeR-BERT$_{base}$ & \textbf{76.22} & \textbf{65.13} & \textbf{52.28} & \textbf{67.31} \\
		\bottomrule
	\end{tabular}
	\caption{Downstream performance on semantic retrieval and reranking datasets, including QQP and AskUbuntu.}
	\label{tab:main_retrieval}
\end{table}

% ubuntu上要不要搞个sbert什

Because sentence representation learning has broad application scenarios, the performance on the STS tasks only is not enough to present the quality of sentence representations. According to TSDAE \cite{wang-etal-2021-tsdae-using}, good STS performance does not necessarily correlate with good performance on downstream semantic retrieval or reranking task, as there exists obvious inductive bias. Therefore, in this section, we conduct extensive experiments on semantic retrieval on the Quora Question Pairs dataset, and semantic reranking on the AskUbuntu dataset. We compare the performance between our PaSeR and other frontier works including SimCSE and BERT-whitening.

Table \ref{tab:main_retrieval} presents the performance of all models on both datasets. Compared to the previous best contrastive method SimCSE, our PaSeR achieves better performance in both supervised and unsupervised settings. Better results on both semantic retrieval and reranking indicate that our PaSeR is better at ranking sentences with similar meanings, which is a core feature that can not be present by STS benchmarks, but is extremely valued in semantic retrieval and reranking.

\subsection{Ablation Study}

\begin{figure}[tp]
    \centering
    \includegraphics[width=1.0\linewidth]{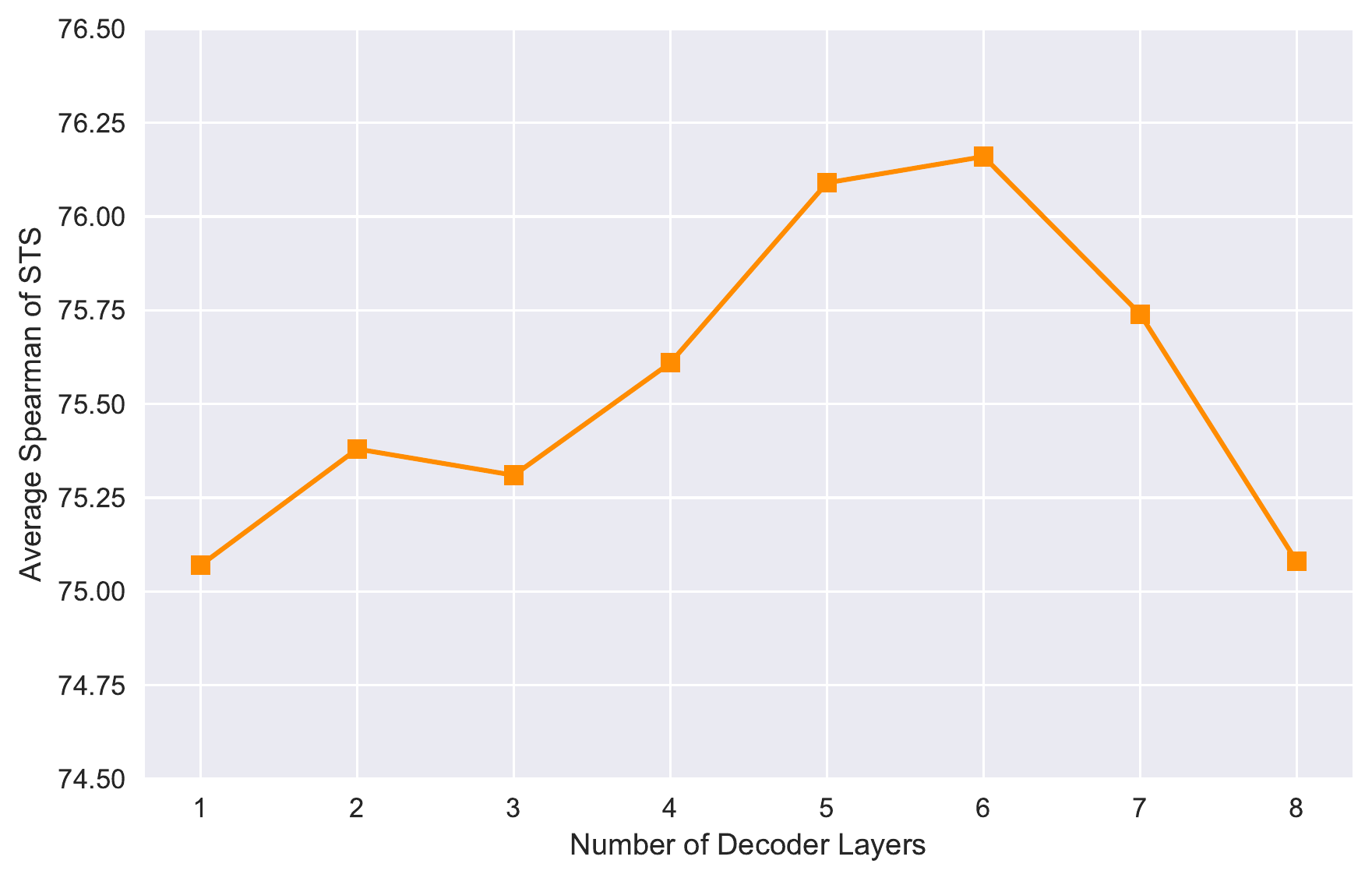}
    \caption{Ablations on the effects of the complexity of the generative decoder.}
    \label{fig:complexity_of_decoder}
\end{figure}

\subsubsection{Complexity of Generative Decoder} \label{discuss:sent_encoder}

Inspired by the design of Electra \cite{clark2019electra}, we further conduct extensive experiments to study the effects brought by the complexity of generative decoders. We vary the layer number of the generative decoder from 1 to 8, and train all versions for 5 epochs on the dataset. The evaluation metric is the average Spearman's correlation of the whole 7 STS tasks. 

The results are shown in Figure \ref{fig:complexity_of_decoder}. From the figure, we can see that the complexity of generative decoders largely affects the performance of the sentence encoder. The sentence encoder achieves its best performance when the layer number is set close to 6. We speculate this is because too small generators lack enough model capacity, while too large generators tend to cause overfitting on the training data.

\begin{table}[tp]
    \centering
    \setlength{\tabcolsep}{5.5pt}
    \fontsize{10.0pt}{\baselineskip}\selectfont
    \begin{tabular}{lccc}
    \toprule
    Model & STS12-16 & STS-B & SICK-R \\ 
    % & Avg.all \\
    \midrule
    BERT-[CLS] & 31.4 & 20.3 & 42.4 \\
    % & 31.4 \\
    \hdashline
    PaSeR$_{base}$-no aug & 73.8 & 74.5 & 62.5 \\ 
    % & 72.3 \\
    
    \ + RS & 74.2 & 75.0 & 63.5 \\
    % & 72.8 \\
    
    \ + RD & 74.2 & 75.5 & 63.8 \\ 
    % & 72.9 \\
    
    \ + SR & \textbf{77.7} & \textbf{79.3} & \textbf{65.4} \\ 
    %& \textbf{75.9} \\
    \ + SR + RD & 77.2 & 77.5 & 64.5 \\
    % & 75.4 \\
    
    \ + SR + RS & 77.3 & 77.6 & 63.1 \\
    % & 75.3 \\
    
    \bottomrule
    \end{tabular}
    \caption{Ablation study of data augmentation strategies, “STS12-16” represents the average Spearman score of STS12-16 datasets.}
    \label{tab:data_aug}
\end{table}

\subsubsection{Effectiveness of Data Augmentation}

In this section, we compare the effects of different data augmentation strategies, shown in Table \ref{tab:data_aug}.

From the experiments, (i) Compared to the original BERT baseline, our PaSeR can already achieve remarkable performance gain without any data augmentation techniques. (ii) Synonymy Replacement (SR) is extremely effective in boosting downstream performance. (iii) Random Swapping (RS) and Random Deletion (RD) also help, but with a much smaller effect. (iv) We do not observe performance gain when different data augmentation techniques are combined. We speculate this is because the semantic meaning of one sentence is more likely to change when different augmentation methods are combined, which influences the alignment of the input sentence pairs.

\subsubsection{Choices of Phrases to Mask} \label{sec:phrase_to_mask}

In this section, we will discuss the effect brought by the masking choices of important phrases. As we have discussed previously, we conduct experiments to examine the two masking strategies we proposed. For masking using NLTK toolkit, we specifically conduct three experiments, including (i) masking out Noun Phrases (NP) only, (ii) masking out Verb Phrases (VP) only, (iii) both NP and VP. For masking using RAKE, we vary the number (from 1 to 5) of the most important phrases we choose to mask in the sentence.

Table \ref{tab:abl_mask_phrase} presents the results of each masking strategy. (i) Apparently, using RAKE as the phrase extraction method achieves significantly better performance than using NLTK toolkit. In fact, the syntax parsing based method views all the phrases in one sentence with equal importance, while it is obvious that different phrases contribute differently to the semantic meaning of one sentence. (ii) For masking using NLTK toolkit, \textit{masking out Noun Phrases only} or \textit{masking out Verb Phrases only} perform worse than \textit{masking both phrases}. We speculate this is because neither Verb Phrases nor Noun Phrases can fully cover the semantic meaning of one sentence. Our PaSeR model needs to encode information of both phrases into the sentence representation. (iii) For masking using RAKE, when compared to the BERT-[CLS] baseline, the top three phrases contribute the most in facilitating the modeling of sentence representation, as adding each one will result in remarkable performance improvement. However, adding more phrases (top 4 or top 5) results in trivial improvement. We speculate this is because the top 4 or 5 phrase contributes little important information to the semantic meaning of one sentence, especially when sentences are often short in the STS benchmark.

\begin{table}[tp]
	\centering
	\setlength{\tabcolsep}{4.0 pt}
	\fontsize{10.0pt}{\baselineskip}\selectfont
	\begin{tabular}{lccc}
		\toprule
		Phrase to Mask & STS12-16 & STS-B & SICK-R \\ 
		%& Avg.all \\
		\midrule
            None (BERT-[CLS]) & 31.4 & 20.3 & 42.4 \\

            \hdashline
		
		NP only & 68.5 & 68.2 & 62.0 \\
		% & 75.0 \\
		
		VP only & 63.8 & 64.1 & 58.3 \\ 
		% & 74.6 \\
		
		NP+VP & 69.5 & 70.4 & 64.3 \\
		% & 75.0 \\
		\hdashline
		RAKE (top1) & 73.5 & 73.9 & 62.4 \\
            RAKE (top2) & 76.8 & 77.6 & 63.6 \\
            RAKE (top3) & \textbf{77.8} & \textbf{79.4} & 64.4 \\
            RAKE (top4) & 77.6 & 79.3 & 64.6 \\
            RAKE (top5) & 77.7 & 79.3 & \textbf{65.4} \\
		% RAKE (random 2)  & \textbf{77.7} & \textbf{79.3} & \textbf{65.4} \\
		% & \textbf{75.9} \\
		
		\bottomrule
	\end{tabular}
	\caption{Ablation study of different phrases to mask in the unsupervised setting. "STS12-16" represents the average Spearman score of STS12-16 datasets.}
	\label{tab:abl_mask_phrase}
\end{table}

\begin{table*}[tp]
    \centering
    \setlength{\tabcolsep}{1.5pt}
    \fontsize{10.0pt}{\baselineskip}\selectfont
    \begin{tabular}{ll}
        \toprule
        \textbf{SimCSE-BERT$_{base}$} & \textbf{PaSeR-BERT$_{base}$}  \\
        \midrule
        \multicolumn{2}{l}{\textbf{Query}: What can one do to relieve severe chronic pain?} \\
        \midrule
        \textit{Is it possible to come to terms with a life of chronic pain?} & \textit{What has worked for you to help relieve chronic pain?} \\
       \textit{What is the best way to avoid pain?} & \textit{How can I get rid of chronic and acute back pain?} \\
        \textit{How would you/do you cope with chronic pain?} & \textit{What is the best way to ease period pain?} \\
        \midrule
        
        \multicolumn{2}{l}{\textbf{Query}: Is there a biological reason that people cry when they are emotional?} \\
        \midrule
        \textit{Why do people cry when they feel happy?} & \textit{Why do some people cry when they get angry?} \\
        \textit{Why do some people cry more than others?} & \textit{Why do people cry when they feel happy? }\\
        \textit{Why do some people like crying so much?} & \textit{Why do some people like crying so much?} \\
        
        % \midrule
        % \multicolumn{3}{l}{\textbf{Query}: \textit{What are natural numbers?}} \\
        % \midrule
        % \#1 & What are real numbers? & What is a natural number? \\
        % \#2 & What are numbers? Do numbers exist? & What is a whole number and a natural number? \\
        % \#3 & How are natural numbers used in everyday life? & What are real numbers? \\
        % \midrule
        \bottomrule
        
    \end{tabular}
    \caption{Qualitative Analysis of semantic retrieval on the QQP Dataset.}
    \label{tab:qualitative_analysis}
\end{table*}

\begin{table*}[tp]
    \centering
    \setlength{\tabcolsep}{8.0pt}
    \fontsize{10.0pt}{\baselineskip}\selectfont
    \begin{tabular}{lll}
        \toprule
        Raw Sentence & Masked version & Generate result \\
        \midrule
        \textit{\textbf{Work} with a tool.} & \textit{\_ with a tool} & \textit{Try} \\
        \textit{At \textbf{least 89 dead} in china earthquakes} & \textit{At \_ \_ \_ dead in china earthquakes} & \textit{least 7 dead.} \\
        \textit{Do I need a transit visa for a stop in \textbf{London}?} & \textit{Do I need a transit visa for a stop in \_?} & \textit{UK} \\
        \textit{There are \textbf{two options} for you} & \textit{There are \_ \_ for you} & \textit{options options} \\
        \bottomrule
        
    \end{tabular}
    \caption{Qualitative Analysis of the phrases that PaSeR decoder reconstructs.}
    \label{tab:generator_perform}
\end{table*}

\section{Qualitative Analysis}
\subsection{Sentence Retrieval}

In this section, we present the qualitative analysis of the retrieval results on Quora Question Pair dataset. We showcase two examples in Table \ref{tab:qualitative_analysis}, where PaSeR retrieves generally better quality sentences. 

In the first case, PaSeR successfully captures the semantic similarity between the phrase \textit{What can one do} and  phrase \textit{What has worked for you} where SimCSE fails. In the second case, PaSeR captures the correlation between \textit{emotional} and \textit{happy/angry}, while SimCSE captures only phrase \textit{happy} in the Top3 prediction. Both cases have demonstrated the superiority of our PaSeR in capturing semantic similar phrases between sentences.

\subsection{Phrases Reconstructed by Decoder}
In this section, we present what the decoder in our PaSeR can do to better illustrate the linguistic interpretability provided by our PaSeR. We sample several sentences, mask one phrase in each of them, and let the decoder reconstruct the missing part. We especially list the cases that our decoder reconstructs differently from the original text in Table \ref{tab:generator_perform} (most of the cases generate the same phrases as original texts).

From the table, we can see that our PaSeR decoder can learn approximately what information is missing from the given sentence representations. It learns phrase similarity between \textit{Work} and \textit{Try}, and the semantic connection between \textit{London} and \textit{UK}. Although it fails to learn the exact arithmetic number \textit{89} and \textit{two}, it is still able to produce a wrong arithmetic number and learn the plurality (\textit{s} in \textit{options}).

\section{Conclusion}
As most pre-trained language models fail to attach enough importance to sentence-level representation learning, it usually leads to unsatisfactory performance in downstream tasks when good sentence representation is right indispensable. Based on investigating the intra-sentence relationship between components of sentences (important phrases) and the whole sentence representations, we propose a generative objective to align these phrases with their corresponding sentence representations. This idea leads to PaSeR, a \textbf{P}hrase-\textbf{a}ware \textbf{Se}ntence \textbf{R}epresentation model. As an effective alternative in Sentence Representation Learning, our PaSeR achieves comparable performance with strong contrastive learning baselines on STS tasks, and better performance on the downstream semantic retrieval and reranking tasks on datasets including QQP and AskUbuntu.

\section{Limitations}
We think our PaSeR has the following limitations, and leave them for future work.

\noindent$\bullet$ The combination of decoding signals is empirically designed. Hyperparameters $m$ and $n$ are selected by grid search and lack technical analysis.

\noindent$\bullet$ From the experiments, what phrases to mask and what augmentations on the sentences are taken can cause significant performance differences. Better masking strategies can be explored.

\bibliography{anthology,custom}
\bibliographystyle{acl_natbib}

\clearpage
\appendix

\section{Hyperparameters for Combined Decoding Signals} \label{sec:dec_signal}

According to SBERT \cite{reimers2019sentence}, the way of combining sentence representations for decoding signal also plays an important role in the downstream performance. In this paper, given the original sentence representation $E_s$ and masked sentence representation $E_{\Tilde{s}}$ we create the decoding signal by the following formula:
\begin{equation}
Signal_{Dec} = [E_s, E_{\Tilde{s}}, m*|E_s - E_{\Tilde{s}}|, n*|E_s * E_{\Tilde{s}}|]   
\end{equation}

In this section, we use grid search to find the best performing $m$ and $n$ that maximize the downstream performance on the STS benchmarks. We first select $m$ from [0.1, 1, 10], and then fix best $m$ to select best $n$ from [0.1, 1, 10]

The results are shown in Table \ref{tab:combine_method}. All experiments use a 6-layer transformer decoder and share the same hyperparameters except for $m$ and $n$. Experimentally, we find that increasing $m$ and $n$ can lead to better performance within the range of [0.1, 10]. Best unsupervised checkpoint is acquired when both $m$ and $n$ are set to 10.

\section{Choices of Pooling Method} \label{sec:pooling_method}
Previous studies \cite{li2020emnlp,su2021whitening,gao2021simcse} have verified that different pooling methods might lead to very different results, and different models may prefer different types of pooling methods. Therefore, we also investigate what pooling method is preferred by our PaSeR. We mainly investigate four types of pooling methods. (i) Average representation over the last layer of BERT, denoted as Top1 avg. (ii) Average representation from the last two layers of BERT, denoted as Top2 avg. (iii) Average representation from the combination of the first layer and the last layer of BERT, denoted as First-last avg. (iv) Directly using the "[CLS]" token as the sentence representation.
\begin{table}[tp]
	\centering
	\setlength{\tabcolsep}{5.5pt}
	\fontsize{10.0pt}{\baselineskip}\selectfont
	\begin{tabular}{lccc}
		\toprule
		Parameter Setting & STS12-16 & STS-B & SICK-R \\ 
		%& Avg.all \\
		\midrule
		
		$m=0.1, n=1.0$ & 77.2 & 78.5 & 64.6 \\
		% & 75.0 \\
		
		$m=1.0, n=1.0$ & 77.6 & 78.7 & 64.8 \\ 
		% & 74.6 \\
		
		$m=10\ ,n=1.0$ & \textbf{77.7} & 79.0 & 65.1 \\
		% & 75.0 \\
		
		$m=10\ ,n=0.1$ & 77.6 & 79.0 & 65.1 \\
		% & \textbf{75.9} \\
		$m=10\ ,n=10$  & \textbf{77.7} & \textbf{79.3} & \textbf{65.4} \\
		\bottomrule
	\end{tabular}
	\caption{Ablation for finding best $m$ and $n$ in the unsupervised setting.}
	\label{tab:combine_method}
\end{table}

Table \ref{tab:abl_pooling} present the results when different pooling methods are applied on PaSeR. Experimentally, we found that directly using "[CLS]" token as the final sentence representation performs the best among all the pooling methods, with nearly 1 point increase on the average performance of all STS tasks.

\begin{table}[tp]
	\centering
	\setlength{\tabcolsep}{8pt}
	\fontsize{10.0pt}{\baselineskip}\selectfont
	\begin{tabular}{lccc}
		\toprule
		Pooling & STS12-16 & STS-B & SICK-R \\ 
		%& Avg.all \\
		\midrule
		
		Top1 avg. & 76.2 & 78.3 & 65.7 \\
		% & 75.0 \\
		
		Top2 avg. & 75.6 & 78.0 & \textbf{66.1} \\ 
		% & 74.6 \\
		
		First-last avg. & 76.2 & \textbf{78.4} & 65.8 \\
		% & 75.0 \\
		
		[CLS]  & \textbf{77.7} & \textbf{79.3} & 65.4 \\
		% & \textbf{75.9} \\
		
		\bottomrule
	\end{tabular}
	\caption{Ablation study of different pooling method on PaSeR.}
	\label{tab:abl_pooling}
\end{table}

\begin{figure*}[tp]
\centering
\subfigure[BERT-{[CLS]}]{
\includegraphics[width=0.30\linewidth]{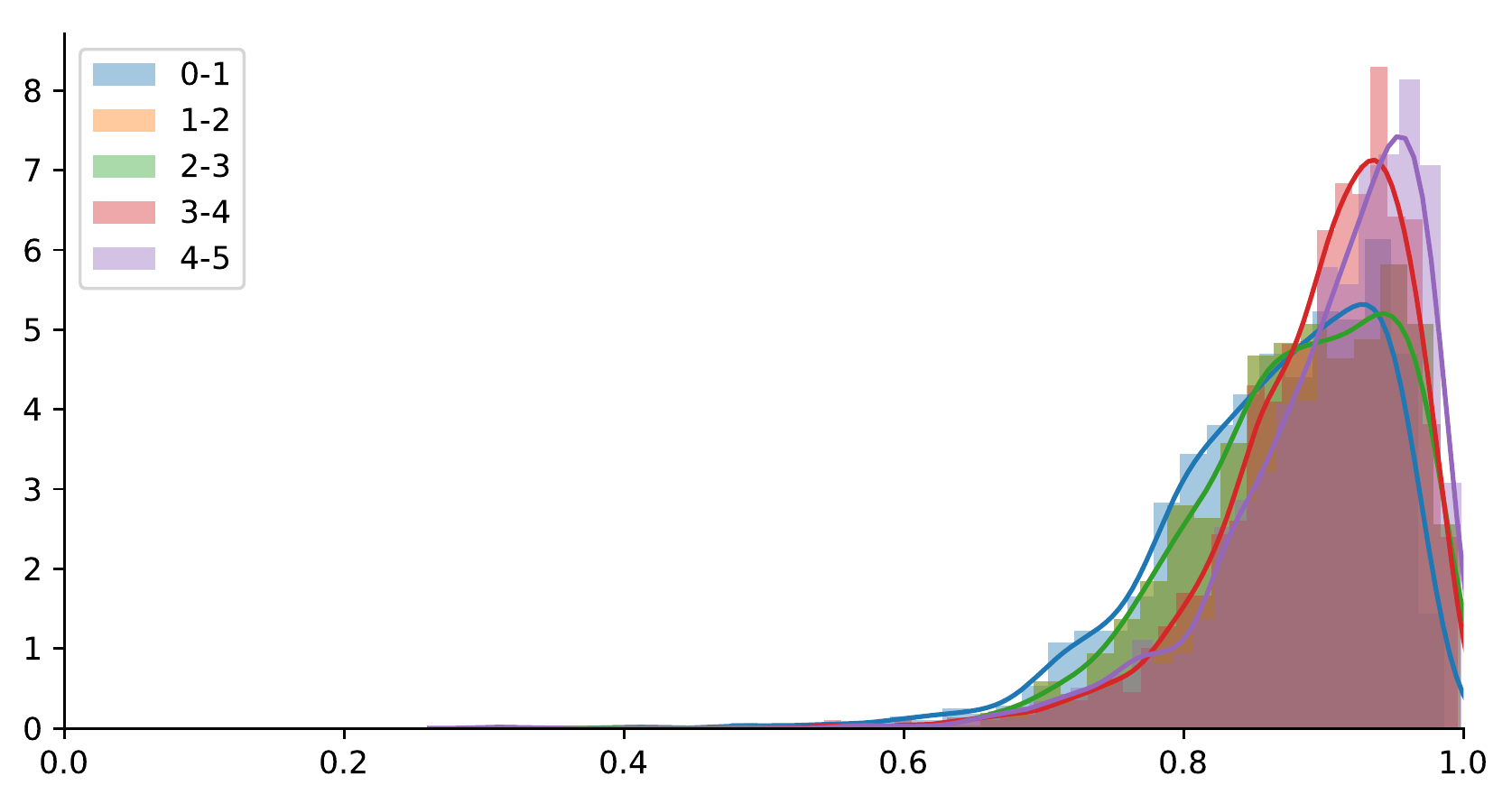}
}
% \quad
% \subfigure[BERT-whitening$_{base}$]{
% \includegraphics[width=0.21\linewidth]{figures/unsup-simcse_density.pdf}
% }
\quad
\subfigure[SimCSE$_{base}$]{
\includegraphics[width=0.30\linewidth]{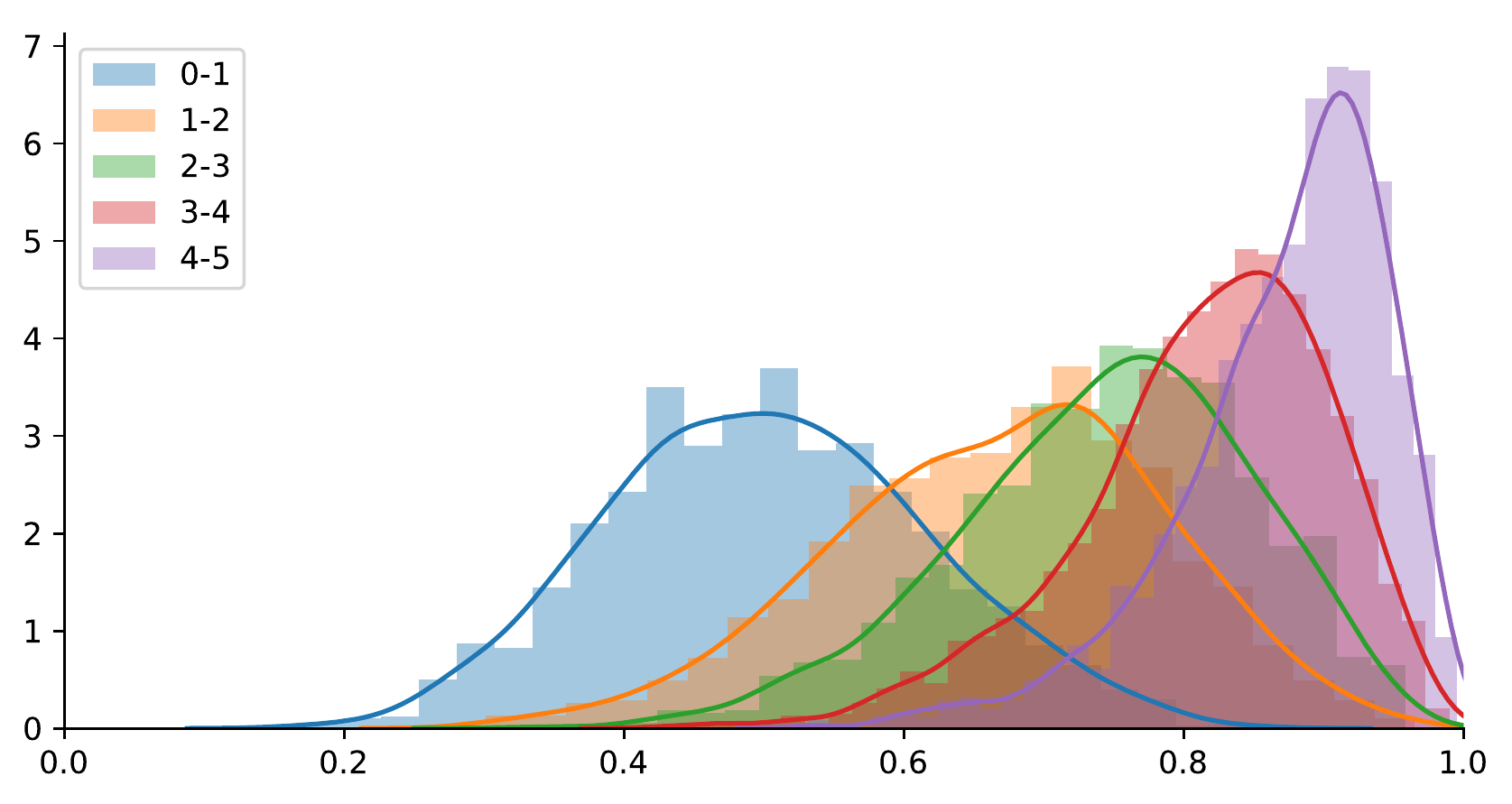}
}
\quad
\subfigure[PaSeR$_{base}$]{
\includegraphics[width=0.30\linewidth]{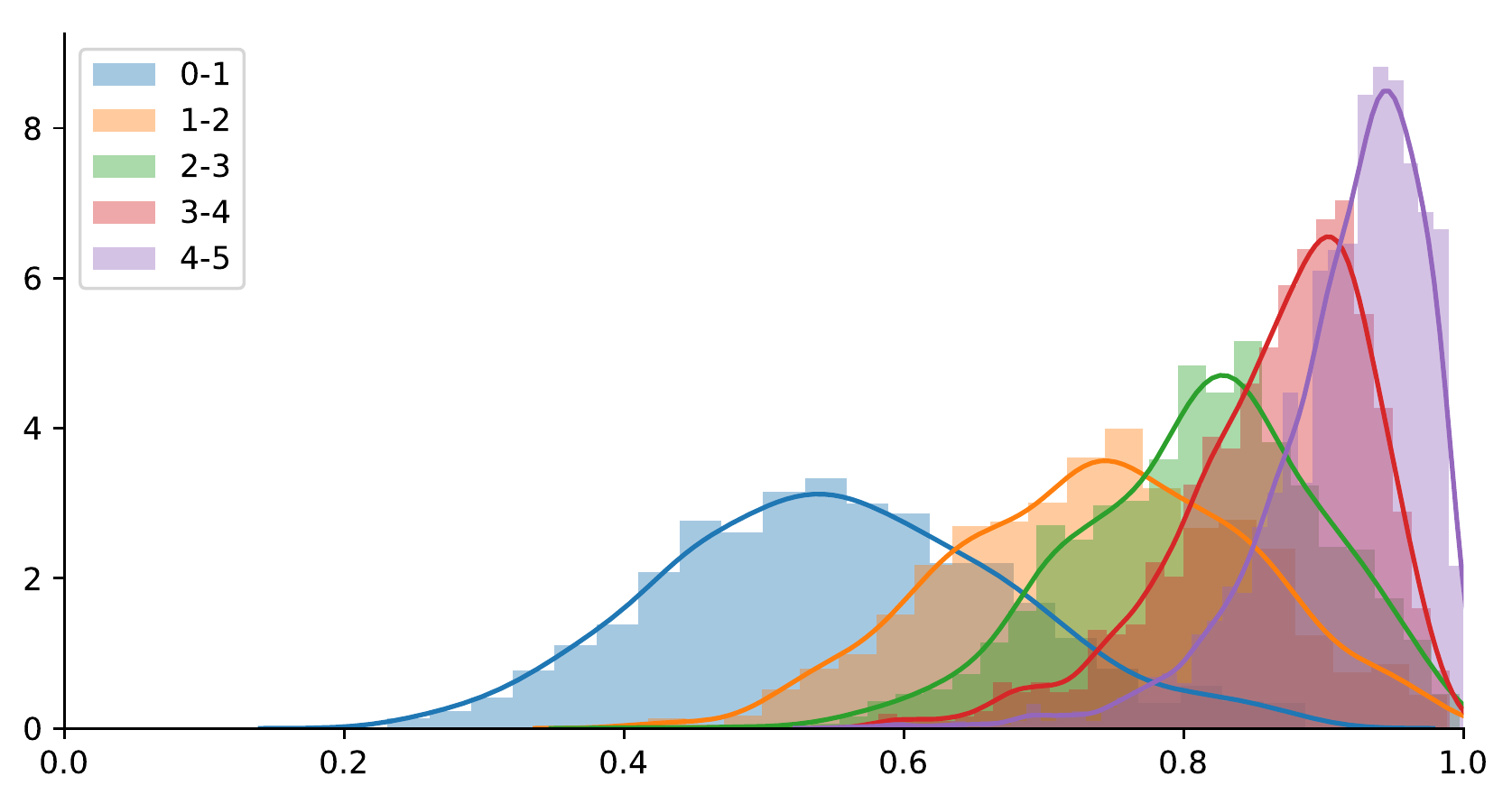}
%\caption{fig1}
}
\caption{Cosine Similarity Density Plots of different models on different similarity levels.}
\label{fig:density_plots}
\end{figure*}

\section{Cosine Similarity Density Plots}

Following SimCSE, we visualize the cosine density plots on the STS-Benchmark dataset in Figure \ref{fig:density_plots}. Concretely, we split the STS-B dataset into five similarity levels according to their labeled scores, and count all similarity scores in each sentence level. From Figure \ref{fig:density_plots}, BERT-[CLS] shows similar cosine distribution in all similarity levels, while SimCSE and PaSeR present good performance in distinguishing samples from different levels.

Theoretically, high-quality sentence representation should present two characteristics. (1) Significant mean value difference between each similarity level, which represents inter-class distance. (2) Lower variance in each similarity level, which represents smaller intra-class distance.

Table \ref{tab:mean_var} presents the exact mean/var values of different models in each similarity level. We can see that both SimCSE and PaSeR achieve good inter-class distance compared to the original BERT-[CLS]. As for intra-class distance, when compared to SimCSE, PaSeR shows generally better performance on almost all similarity levels.

\begin{table}[tp]
    \centering
    \fontsize{10.0pt}{\baselineskip}\selectfont
    \setlength{\tabcolsep}{6pt}
    \begin{tabular}{cccc}
        \toprule
        level & BERT-[CLS] & SimCSE$_{base}$ & PaSeR$_{base}$ \\
        \midrule
        0-1 & 0.86\ /\ 0.006 & 0.50\ /\ 0.013 & 0.55\ /\ 0.015 \\
        1-2 & 0.88\ /\ 0.006 & 0.67\ /\ 0.014 & 0.74\ /\ 0.011 \\
        2-3 & 0.89\ /\ 0.006 & 0.74\ /\ 0.010 & 0.81\ /\ 0.008 \\ 
        3-4 & 0.90\ /\ 0.004 & 0.81\ /\ 0.008 & 0.86\ /\ 0.005 \\
        4-5 & 0.90\ /\ 0.005 & 0.87\ /\ 0.006 & 0.92\ /\ 0.003 \\
        \bottomrule
    \end{tabular}
    \caption{Mean/variance of cosine similarity on STS-Benchmark.}
    \label{tab:mean_var}
\end{table}

\section{Uniformity and Alignment}\label{sec:uniform_align}
% 那这个我们可以去分析看看了。
\begin{figure}[tp]
    \centering
    \includegraphics[width=1.0\linewidth]{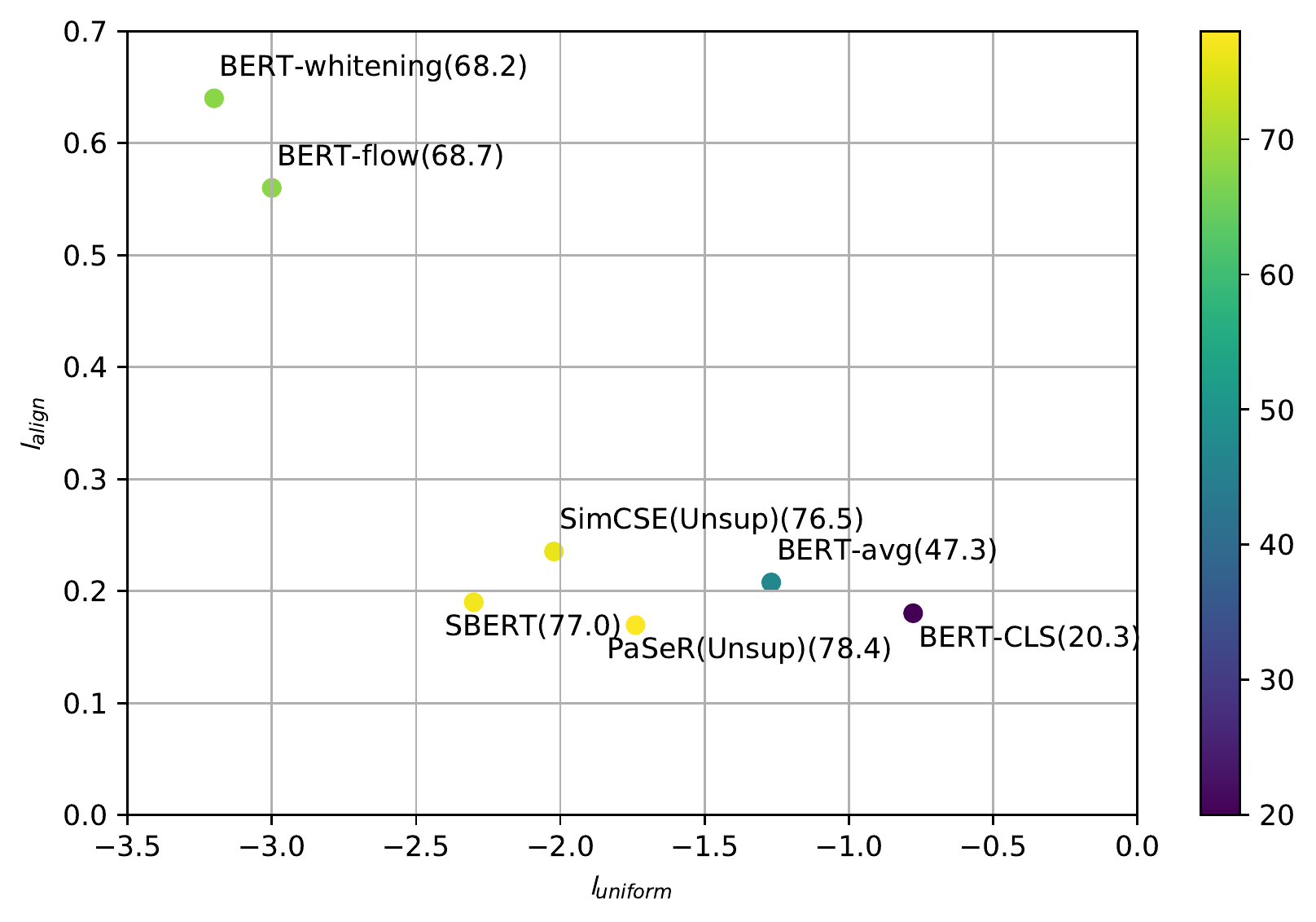}
    \caption{Visualization of uniformity and alignment for sentence representations produced by different methods. All models are trained on BERT$_{base}$. Color of points and numbers in brackets represent Spearman's correlation on the test set of STS-Benchmark.}
    \label{fig:uniform}
\end{figure}

Following SimCSE \cite{gao2021simcse}, we analyze the uniformity and alignment \cite{wang2020understanding} of our PaSeR along with other frontier works. For "uniformity", the representations of all sentences should be approximately uniformly distributed on a unit hypersphere, in order to preserve as much information as possible. While for "alignment", similar sentences should have similar representations. We use the STS-Benchmark dataset as the evaluation corpus, and also list the corresponding Spearman correlation for each method for comparison.

From Figure \ref{fig:uniform}, we can see that our PaSeR achieves the best alignment loss among all the listed models (0.17), which is even better than supervised baseline SBERT (0.19) or the SOTA unsupervised method SimCSE (0.24). For the uniformity measurement, previous works \cite{li2020emnlp,su2021whitening} have pointed out that original BERT sentence representation space \textit{collapse}, which presents high similarity between representations of any sentence pairs. Therefore, both BERT-avg and BERT-[CLS] suffer from high uniformity loss. When compared to BERT-[CLS] or BERT-avg, our PaSeR also achieves much better uniformity, meaning that our proposed self-supervised sentence-level training objective naturally eases the \textit{collapse}.

\end{document}